\documentclass[letterpaper]{article} 
\usepackage{aaai2026}  
\usepackage{times}  
\usepackage{helvet}  
\usepackage{courier}  
\usepackage[hyphens]{url}  
\usepackage{graphicx} 
\usepackage{natbib}  
\usepackage{caption} 
\usepackage{algorithm}
\usepackage{algorithmic}
\usepackage{booktabs}

\usepackage{amsmath}
\usepackage{newfloat}
\usepackage{listings}
\DeclareCaptionStyle{ruled}{labelfont=normalfont,labelsep=colon,strut=off} 
\floatstyle{ruled}
\newfloat{listing}{tb}{lst}{}
\floatname{listing}{Listing}
\title{FGM-HD: Boosting Generation Diversity of Fractal Generative Models through Hausdorff Dimension Induction}

\author {
    Haowei Zhang\textsuperscript{1,2}, 
    Yuanpei Zhao\textsuperscript{1,2}, 
    Ji-Zhe Zhou\textsuperscript{1,2}\thanks{Corresponding authors: Ji-Zhe Zhou, Mao Li} , 
    Mao Li\textsuperscript{1,2}
}
\affiliations {
    \textsuperscript{1}College of Computer Science, Sichuan University, China\\
    \textsuperscript{2}Engineering Research Center of Machine Learning and Industry Inteligence, Ministry of Education of China\\
    \{zhanghaowei1, zhaoyuanpei\}@stu.scu.edu.cn, \{jzzhou, limao\}@scu.edu.cn
}

\usepackage{multirow} 

\usepackage{bibentry}

\begin{document}

\maketitle

\begin{abstract}

Improving the diversity of generated results while maintaining high visual quality remains a significant challenge in image generation tasks. Fractal Generative Models (FGMs) are efficient in generating high-quality images, but their inherent self-similarity limits the diversity of output images. To address this issue, we propose a novel approach based on the Hausdorff Dimension (HD), a widely recognized concept in fractal geometry used to quantify structural complexity, which aids in enhancing the diversity of generated outputs. To incorporate HD into FGM, we propose a learnable HD estimation method that predicts HD directly from image embeddings, addressing computational cost concerns. However, simply introducing HD into a hybrid loss is insufficient to enhance diversity in FGMs due to: 1) degradation of image quality, and 2) limited improvement in generation diversity. To this end, during training, we adopt an HD-based loss with a monotonic momentum-driven scheduling strategy to progressively optimize the hyperparameters, obtaining optimal diversity without sacrificing visual quality. Moreover, during inference, we employ HD-guided rejection sampling to select geometrically richer outputs. Extensive experiments on the ImageNet dataset demonstrate that our FGM-HD framework yields a 39\% improvement in output diversity compared to vanilla FGMs, while preserving comparable image quality. To our knowledge, this is the very first work introducing HD into FGM. Our method effectively enhances the diversity of generated outputs while offering a principled theoretical contribution to FGM development.

\end{abstract}


\section{Introduction}

Generative models have achieved remarkable progress in recent years, particularly in the image synthesis domain. Approaches such as generative adversarial networks (GANs)~\cite{goodfellow2014generative, wiatrak2019stabilizing}, variational autoencoders (VAEs)~\cite{kingma2013auto, van2017neural}, diffusion models~\cite{ho2020denoising, rombach2022high, shen2025efficient}, and autoregressive and flow-based models~\cite{van2016pixel, kingma2018glow, you2022locally}, have demonstrated the ability to produce high-fidelity and semantically coherent images. However, maintaining an adequate balance between image \textbf{quality} and \textbf{diversity} remains a fundamental challenge. Although many models excel at generating visually appealing outputs, they often fail to capture the full variability of the underlying data distribution.

Fractal Generative Models (FGMs)~\cite{li2025fractal} offer a unique architecture for high-quality image generation by leveraging recursive self-similarity, an intrinsic property of fractals. By repeatedly applying a compact generative module across multiple scales, FGMs can efficiently generate complex, globally consistent, and high-resolution images. This modular recursive design makes FGMs particularly suitable for tasks requiring structural coherence and visual richness. Nevertheless, the same self-similar structure that ensures global consistency can also result in repetitive patterns and insufficient diversity in generated results.

To address the challenge of limited diversity in FGMs, we introduce the \textbf{Hausdorff Dimension (HD)}~\cite{hausdorff1918dimension} as a geometric indicator for structural complexity. As a numerical concept in fractal geometry, HD quantifies the variation of spatial detail across scales, with higher values typically reflecting greater structural richness. This makes HD particularly suitable for guiding the generation of diverse and intricate outputs.

Nevertheless, conventional HD estimation techniques, such as the box counting method~\cite{mandelbrot1983fractal}, are computationally expensive and not easily compatible with large-scale training pipelines. In addition, HD measures structural complexity, and deep layer embeddings learned by neural networks have been proven to capture this structure effectively~\cite{valle2022characterization, werbos2002backpropagation}. Therefore, at the commencement, we propose a learning-based HD estimation method for efficient HD prediction directly from image embeddings to reduce computational costs while maintaining high accuracy, making it feasible for integration into training pipelines.

However, simply introducing HD into a hybrid loss function is insufficient for improving generation diversity, as it leads to degraded image quality and limited improvement in generation diversity. We observed that in the hybrid loss function, the relative weighting of each component is critical: during the early training phase, image quality is typically suboptimal and HD estimates are unreliable, necessitating a focus on quality while gradually increasing the weight of diversity as training progresses, without sacrificing visual quality. Therefore, an optimal balance between HD loss and generative loss is a key factor for effective training. To address this, we propose the \textbf{Monotonic Momentum-Driven Scheduling (MMDS)} strategy, which dynamically adjusts the influence of HD loss over time. The MMDS strategy ensures that the model first focuses on high-quality structure and progressively incorporates diversity, providing a smoother transition and more stable optimization without compromising generation quality.

To further enhance output diversity during inference, we introduce an HD-guided sampling strategy. Leveraging the recursive structure of FGMs, which naturally supports the generation of multiple candidate patches in parallel, we retain only those outputs whose estimated HD exceeds a predefined threshold. This post hoc selection process filters out structurally simple images, resulting in a final output set with greater geometric richness and perceptual variety, achieved without modifying the underlying generative architecture.

Experiments on the ImageNet dataset indicate that our approach significantly improves diversity while maintaining visual quality. In particular, our method achieves a 39\% improvement in Recall compared to the vanilla FGM, highlighting HD's effectiveness as both a training signal and a sampling criterion. To our knowledge, this is the first systematic effort to introduce HD into fractal generative modeling. Additionally, the MMDS strategy for dynamic weight optimization offers a generalizable framework that can be applied to other models utilizing hybrid loss functions, thereby extending its applicability across a broader range of generative tasks.

In summary, our main contributions are as follows.

\begin{itemize}
    \item \textbf{Introducing Hausdorff Dimension for Diversity Enhancement:} We are the first to introduce HD into the FGM framework, proposing a set of methods during both training and inference, including the MMDS and Sampling Strategy, to enhance generation diversity while maintaining high image quality. Our work alleviates the self-similarity limitation inherent in fractal-based generation.
    \item \textbf{Developing a learnable and efficient HD estimation module:} To overcome the inefficiency of numerical HD estimation methods, we design a learnable HD predictor that operates directly on image embeddings, enabling fast and accurate HD prediction and supporting scalable integration into generative frameworks.
    \item \textbf{Monotonic Momentum-Driven Scheduling (MMDS) strategy:} We introduce MMDS, a strategy that dynamically adjusts the influence of HD loss during training to balance visual quality and structural diversity, ensuring stable optimization. MMDS can also be extended to models with hybrid loss functions, providing a systematic approach to optimize the balance between loss components in diverse generative tasks.

\end{itemize}

\begin{figure}[t]
    \centering
    \includegraphics[width=\linewidth]{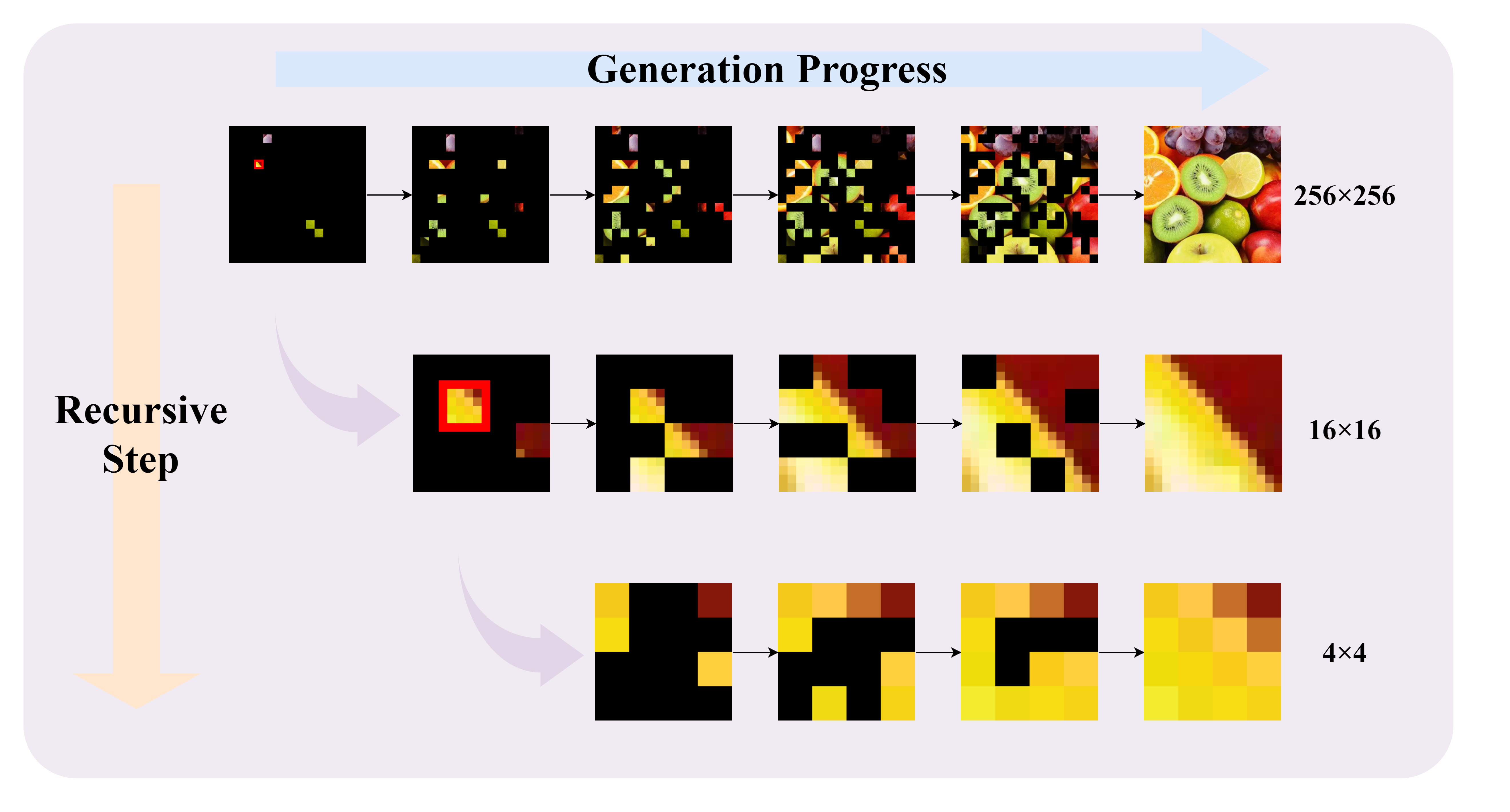}
    \caption{Overview of the FGM process, where the image is generated from sparse patches to $256 \times 256$ resolution through recursive refinement at smaller scales (e.g., $16 \times 16$ and $4 \times 4$ blocks). A shared generative module is reused across scales, capturing global and fine-grained structural details.}
    \label{fig:fgm_structure}
\end{figure}

\section{Related Work}
\subsection{Generative Models}

Generative models have made significant strides in image generation tasks. For instance, \textit{GANs}~\cite{goodfellow2014generative} have enabled high-quality image generation through adversarial learning and have been widely applied in various tasks. Similarly, \textit{VAEs}~\cite{kingma2013auto} leverage probabilistic modeling for generating images through latent variable distributions. More recently, \textit{diffusion models}~\cite{ho2020denoising, rombach2022high, shen2025efficient} have gained popularity for generating high-quality and diverse images by iteratively denoising samples. Moreover, \textit{autoregressive and flow-based models}~\cite{van2016pixel, kingma2018glow, you2022locally} excel in generating high-quality images with strong mode coverage. Despite these advancements, a key challenge remains achieving a balance between image quality and diversity~\cite{lan2024generative, li2021combination}, as existing models struggle to generate sufficiently varied samples that comprehensively reflect the underlying data distribution.

\begin{figure*}[t]
    \centering
    \includegraphics[width=\textwidth]{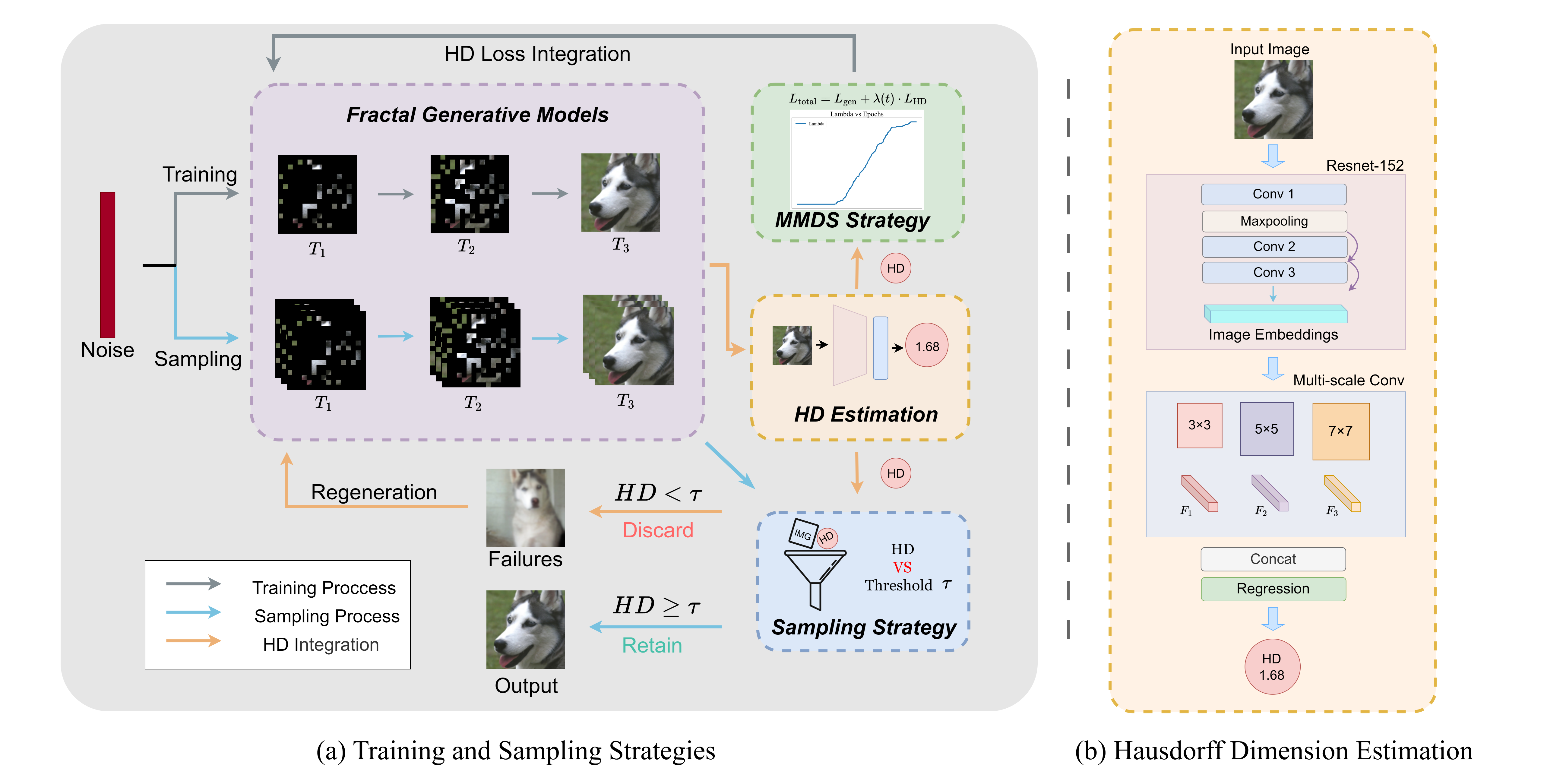} 
    \caption{Overview of the proposed FGM-HD framework. (a) \textbf{Training and Sampling Strategies:} During training (gray line), input noise is recursively processed by the FGM (purple section), and the generated images are evaluated by the HD estimation module to compute HD loss. Then, the HD loss is dynamically weighted by the MMDS strategy (green section) through $\lambda(t)$ to balance image quality and structural diversity during optimization. During inference (blue line), a batch of samples is generated from noise via FGM and passed through the Sampling Strategy (blue section). Only geometrically richer outputs with HD values above the threshold ($ \tau $) are retained while others are discarded and regenerated, ensuring structurally diverse outputs without modifying the generator architecture. (b) \textbf{Hausdorff Dimension Estimation:} The HD estimation (yellow section) is performed using a multi-scale convolutional network built upon the ResNet152 architecture, enabling accurate and efficient HD prediction directly from image embeddings.}
    \label{fig:all}
\end{figure*}

\subsection{Fractal Generative Models}  
Fractal-based generative models leverage the recursive and self-similar properties of fractals to synthesize complex visual patterns. Classical approaches such as iterated function systems (IFS)~\cite{barnsley2014fractals} and Lindenmayer systems (L-systems)~\cite{prusinkiewicz2012algorithmic} generate intricate structures through predefined rules, offering mathematical simplicity but lacking adaptability. Recent advances have incorporated neural networks into fractal frameworks, notably \textit{FGMs}~\cite{li2025fractal}, employing recursive atomic modules to progressively refine images across multiple scales. This design enables FGMs to efficiently generate high-quality and visually coherent images with minimal architectural complexity.

Although FGMs capture structural detail and visual quality, their recursive nature limits output diversity, leading to redundancy and hindering the representation of complex data distributions. Thus, enhancing diversity remains a critical challenge for improvement.

\subsection{Hausdorff Dimension}
The HD concept~\cite{hausdorff1918dimension} was originally introduced to measure the complexity of geometric sets, enabling a precise characterization of self-similar or irregular structures. For instance, ~\citet{khrulkov2019universality} showed that generative models can approximate data supported on low-dimensional manifolds, with HD providing a useful measure of their expressive capacity. ~\citet{simsekli2020hausdorff} demonstrated that HD serves as a proxy for model generalization, revealing a strong correlation between geometric complexity and generalization ability. More recently, ~\citet{li2021hausdorff} proposed Hausdorff GAN, which incorporates HD into the training objective to align the intrinsic dimensionality of real and generated data, resulting in improved output quality and diversity. Leveraging HD, FGMs can generate more diverse outputs while maintaining structural integrity, addressing the limitations of traditional fractal self-similarity.


\section{Methods}
\subsection{Preliminary of Fractal Generative Models (FGMs)}

 Inspired by the recursive and self-similar nature of fractals, FGMs generate complex patterns by repeatedly applying the same generative module at different scales, enabling them to capture the data distribution's global and local features. This recursive design allows FGMs to generate high-quality images efficiently while maintaining multiscale structural details, making them suitable for real-time inference and deployment on resource-constrained devices. An overview of the generation process is illustrated in Figure~\ref{fig:fgm_structure} consisting of the following steps:
 

\begin{itemize}
    \item \textbf{Initial Generation.}  
    First, a low-resolution image is generated using a base generative model. We use the masked autoregressive model (MAR)~\cite{li2024autoregressive} for continuous value autoregression, employing a diffusion-based loss function to avoid discrete tokenization. This serves as the foundation for further refinement.
    
    \item \textbf{Recursive Refinement.}  
    The image is progressively refined through the recursive application of the generative module. Initially, the image is divided into coarse regions, which are subdivided further at each level. Then, the module synthesizes higher-resolution content, capturing global structure and local details. At level $l$, the image $I_l$ is generated from $I_{l-1}$ as follows:
    \begin{equation}
        I_l = {Module}(I_{l-1}, \theta_l),
    \end{equation}
    where $\theta_l$ represents the parameters of the generative module at level $l$.

    \item \textbf{Final Output.}  
    After several recursive steps, a high-resolution image is produced, characterized by intricate details that emerge through the recursive process.
\end{itemize}

\subsection{Hausdorff Dimension Estimation}

The widely used \textit{box counting method}~\cite{mandelbrot1983fractal} for HD estimation is accurate but computationally expensive for high-resolution images and sensitive to noise and edge complexities, limiting its applicability for modern image analysis~\cite{gneiting2012estimators}. HD reflects an image's structural features, with intricate textures and self-similar patterns generally exhibiting higher values~\cite{napolitano2012fractal}. \textit{Convolutional neural networks (CNNs)}\cite{lecun1989backpropagation}, due to learning hierarchical representations, are well-suited to capture these structural features for efficient HD estimation~\cite{valle2022characterization}.

To address the limitations of the box counting method and leverage CNNs, we propose a novel method that directly extracts dimensionality features from image embeddings. Our approach utilizes the \textit{ResNet152}~\cite{he2016deep} architecture, enhanced with a \textbf{multi-scale convolutional module}, as shown in Figure~\ref{fig:all} (b). This model leverages the deep feature extraction capabilities of ResNet152, enhanced by multi-scale convolutions, to capture image features across different scales, which is then used in a regression layer for efficient and accurate HD prediction. Specifically, our method achieves superior HD estimation precision while significantly reducing computational overhead, making it more suitable for real-world applications. 


\section{Framework}

\subsection{Training Strategy}

To balance image quality and structural diversity during training, we formulated the learning objective as a hybrid loss function that combines the base generative loss and the \textit{Hausdorff Dimension loss (HD loss)}, which is defined as:

\begin{equation}
L_{\text{HD}} = \left| \text{HD}{_\text{gen}} - \text{HD}{_\text{target}} \right|,
\end{equation}

where $\text{HD}{_\text{gen}}$ denotes the estimated HD of a generated image, and $\text{HD}{_\text{target}}$ represents the class-specific median HD from the training dataset.

To incorporate HD loss into training in a stable and effective manner, we introduced a novel scheduling strategy termed \textbf{Monotonic Momentum-Driven Scheduling (MMDS)}, which defines a time-dependent weighting function $\lambda(t)$ that gradually increases during training to control the influence of HD loss. The overall training objective becomes:

\begin{equation}
L_{\text{total}} = L_{\text{gen}} + \lambda(t) \cdot L_{\text{HD}},
\end{equation}

where $L_{\text{gen}}$ is the primary generative loss, and $\lambda(t)$ is a non-negative, monotonically increasing coefficient.

\begin{figure}[t]
    \centering
    \includegraphics[width=\linewidth]{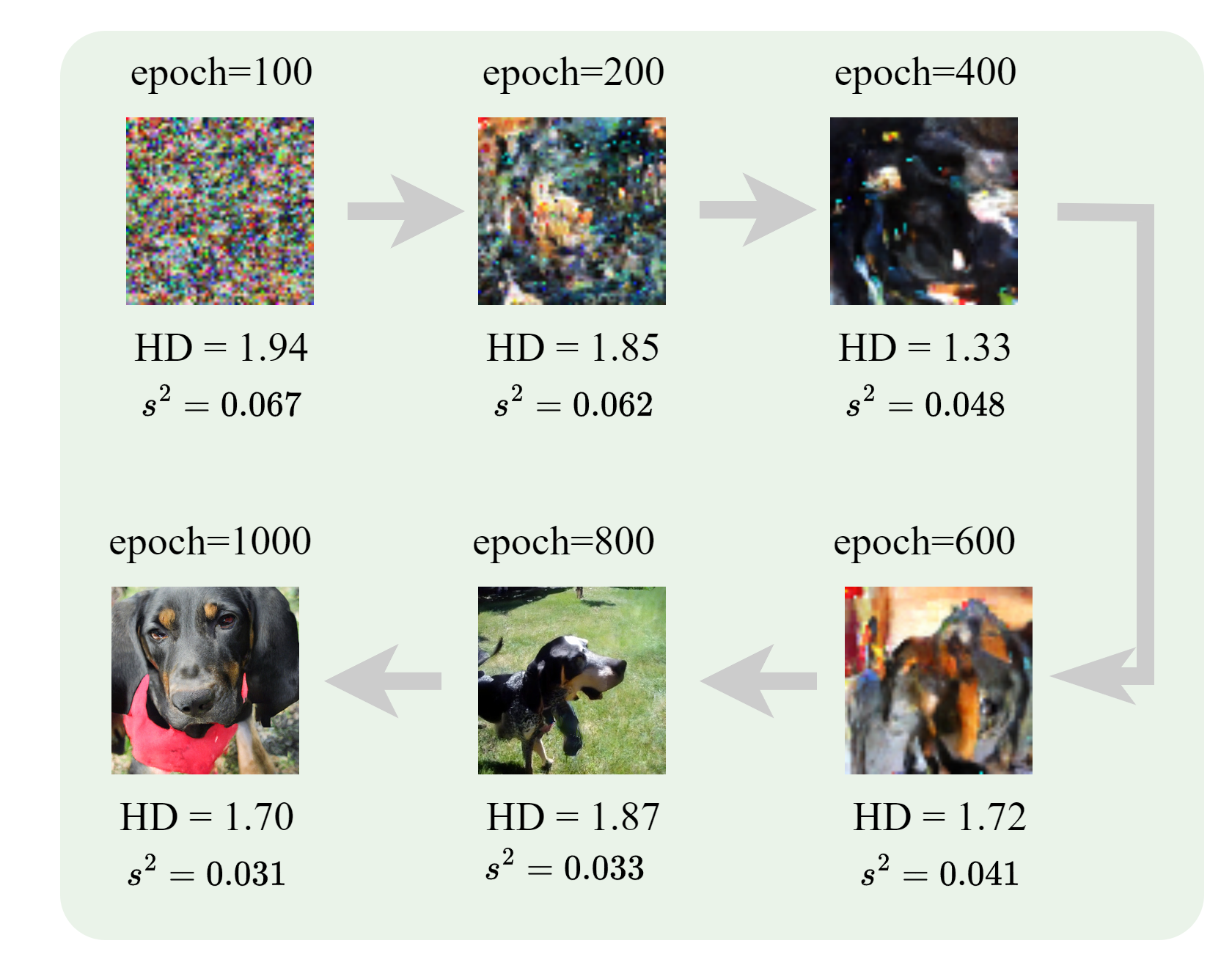}
    \caption{Evolution of image quality and HD variance across training epochs. Early-stage generations are noisy with unstable HD values, while later epochs yield high-quality images with more reliable HD estimation.}
    \label{fig:epoch_with_image}
\end{figure}

As shown in Figure~\ref{fig:epoch_with_image}, in the early stages of training, the model generates noisy images, leading to HD estimates with large deviations and high variance, making HD estimation highly inaccurate and unstable. As training progresses, with the generated images improved, HD estimation becomes more reliable and it is feasible to gradually introduce HD loss to enhance structural diversity without compromising image quality. To implement this progression, $\lambda(t)$ is constructed via a momentum-driven accumulation scheme, inspired by the SGD~\cite{robbins1951stochastic}. This scheme ensures that $\lambda(t)$ increases monotonically, with its rate of growth controlled by the accumulated momentum. This momentum-driven approach reflects the training state, allowing for smooth and controlled growth over time, which avoids abrupt changes in HD loss weighting, outlined in Algorithm~\ref{alg:mmds}. Compared to static schedules, such as exponential or linear increases, MMDS is better suited to dynamically adjust to the model’s evolving requirements during training. As demonstrated in our experiments in Figure~\ref{fig:lambda_loss_plot}, MMDS results in a smoother loss trajectory and improved convergence stability while effectively balancing image quality and structural diversity throughout training.

\begin{algorithm}[t]
\caption{Monotonic Momentum-Driven Scheduling (MMDS) Strategy}
\label{alg:mmds}
\begin{algorithmic}[1]
\REQUIRE Epochs $E$, momentum $\mu \in [0,1]$, scale factor $\gamma$
\STATE Initialize $\lambda \gets 0$, $m \gets 0$, $L_{\text{prev}} \gets 0$
\FOR{epoch $e$ to $E$}
    \STATE Compute validation loss $L_{\text{val}}$
    \STATE $\Delta L \gets \max(0, L_{\text{prev}} - L_{\text{val}})$
    \STATE $m \gets \mu \cdot m + (1 - \mu) \cdot \gamma \cdot \Delta L$
    \STATE $\lambda \gets \lambda + m$  
    \STATE $L_{\text{prev}} \gets L_{\text{val}}$
    \STATE Compute $L_{\text{total}} = L_{\text{gen}} + \lambda \cdot L_{\text{HD}}$
    \STATE Backpropagate and update model
\ENDFOR
\end{algorithmic}
\end{algorithm}

\subsection{Sampling Strategy}
To further promote structural diversity in FGM's outputs, we introduced a post-processing strategy termed \textit{Hausdorff Dimension sampling (HD sampling)}. This method operates after initial sample generation and selectively retains images that exhibit sufficient structural complexity, as measured by their estimated HD. The complete filtering procedure is outlined in Algorithm~\ref{alg:hd-sampling}.

Specifically, once a batch of candidate images is generated, we computed each sample's HD and compared it with a predefined threshold. Only samples with HD values exceeding this threshold are retained, while others are discarded and regenerated from the beginning, shown in Figure~\ref{fig:all} (a).

By filtering out low-HD images, this method effectively suppresses overly smooth, repetitive, or degenerate outputs, resulting in a final sample set that is visually diverse and structurally rich. As this process is applied post-generation, it incurs no additional computational cost during training and remains fully compatible with any pre-trained generative model.

\begin{algorithm}[t]
\caption{HD Sampling Strategy}
\label{alg:hd-sampling}
\begin{algorithmic}[1]
\REQUIRE Threshold $T_{\text{HD}}$, batch of generated samples $\{I_1, I_2, ..., I_N\}$
\STATE Initialize empty list of selected samples $S$
\FOR{each generated sample $I_i$ in $\{I_1, I_2, ..., I_N\}$}
    \STATE Compute $\text{HD}_{\text{generated}}(I_i)$ \hfill 
    \IF{$\text{HD}_{\text{generated}}(I_i) \geq T_{\text{HD}}$}
        \STATE Add $I_i$ to selected list $S$
    \ELSE
        \STATE Regenerate $I_i$ from scratch \hfill
    \ENDIF
\ENDFOR
\STATE \textbf{Return} selected list $S$ \hfill
\end{algorithmic}
\end{algorithm}


\section{Experiments}

To validate our proposed FGM-HD framework's effectiveness, we conducted the following experiments: a) evaluating the performance of HD estimation network, b) assessing the overall effectiveness of the FGM-HD framework. All experiments were conducted on NVIDIA RTX 4090 GPUs, using the PyTorch framework for deep learning.


\subsection{Dataset}

\textbf{Fractal Dataset}. First, we constructed a dedicated dataset to systematically study the HD behavior in controlled settings. This dataset consists of 1,200 images evenly divided into three categories: (i) a collection of canonical fractal images (e.g., Sierpinski triangle and Koch snowflake) with known theoretical HD; (ii) a diverse set of fractal patterns generated using IFS~\cite{barnsley2014fractals} with varying parameters to create a wide range of structural complexities; and (iii) FractalDB images~\cite{kataoka2020pre} with HD values that are approximated using the box counting method. This balanced composition enables the validation of HD estimation accuracy and benchmarking of generative models on data with clearly defined fractal complexity.

\noindent \textbf{Generation Dataset}. Additionally, we evaluated our HD-based methods on the ImageNet-1000 dataset~\cite{deng2009imagenet}, which contains high-resolution images from 1,000 diverse categories, and all the images are resized to $256 \times 256$ for consistency. Using our HD estimation network, we analyzed structural complexity across categories and applied class-wise HD targets during training. The images generated by our proposed \textbf{FGM-HD} are compared with outputs from publicly available pre-trained generative models.

\begin{table}[t]
    \centering
    \renewcommand{\arraystretch}{1.1}
    \setlength{\tabcolsep}{5pt}  
    \begin{tabular}{l|c|c|c}
        \toprule
        \textbf{Method} & \textbf{Type} & \textbf{Error $\downarrow$} & \textbf{Time(s) $\downarrow$} \\
        \midrule
        Box Counting & \multirow{4}{*}{\centering Non-learning} & \textbf{0.002} & 4.70 \\
        Power Spectrum & & 0.079 & 3.41 \\
        Perimeter-Area & & 0.137 & 6.85 \\
        Sandbox Method & & 0.094 & 5.73 \\
        \midrule
        ResNet152 & \multirow{2}{*}{\centering Learning} & 0.012 & 0.40 \\
        \textbf{Ours} & & 0.005 & \textbf{0.32} \\
        \bottomrule
    \end{tabular}
    \caption{Comparison of HD estimation methods, along with their accuracy (Error) and runtime (Time).}
    \label{tab:hd-method-comparison}
\end{table}

\begin{figure}[t]
    \centering
    \includegraphics[width=\columnwidth]{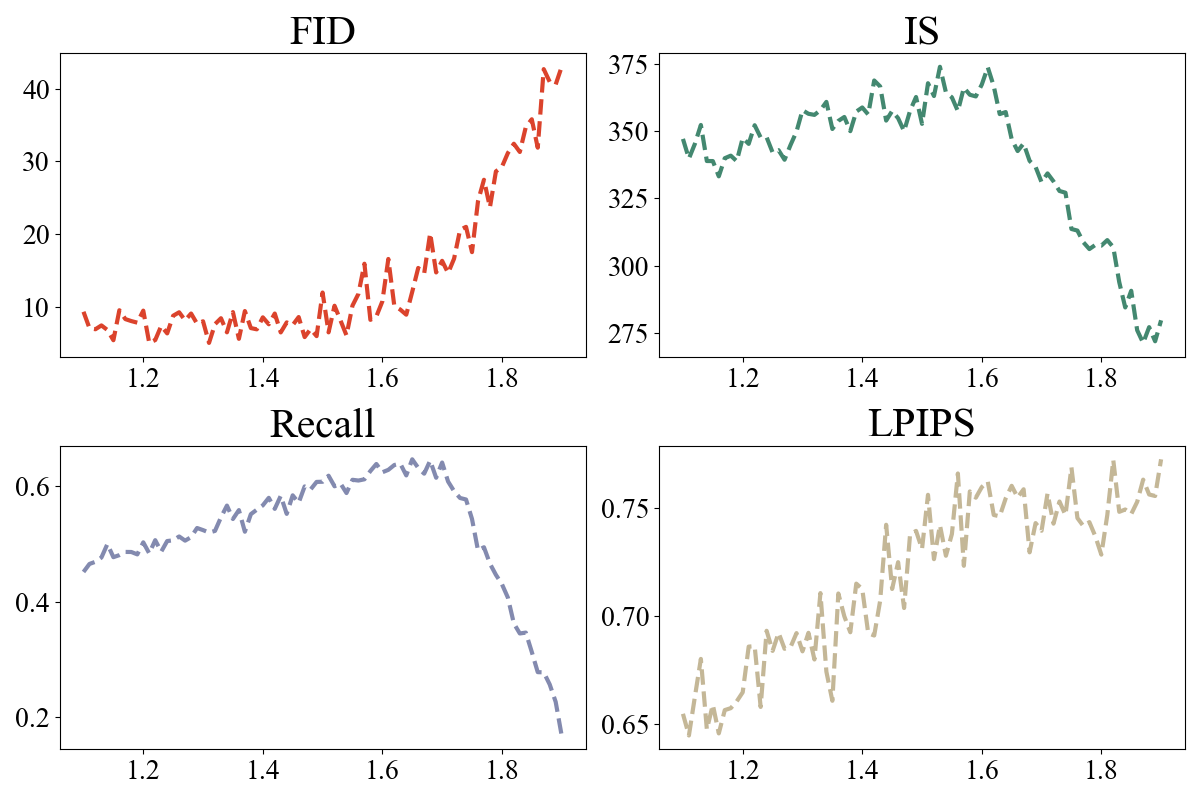}
    \caption{Performance trends of evaluation metrics under varying HD thresholds. }
    \label{fig:metrics_hd_combined}
\end{figure}

\begin{table}[t]
    \centering
    \renewcommand{\arraystretch}{1.1}
    \setlength{\tabcolsep}{5pt}  
    \small
    \begin{tabular}{l|c|c|c|c}
        \toprule
        \textbf{Model} & \textbf{Type} & \textbf{FID} $\downarrow$ & \textbf{IS} $\uparrow$ & \textbf{Recall} $\uparrow$ \\
        \midrule
        BigGAN-deep    & \multirow{3}{*}{GAN}        & 6.95 & 198.2 & 0.28 \\
        GigaGAN        &                           & 3.45 & 225.5 & 0.61 \\
        StyleGAN-XL    &                           & 2.30 & 265.1 & 0.53 \\
        \midrule
        ADM            & \multirow{5}{*}{Diffusion}  & 4.59 & 186.7 & 0.52 \\
        Simple Diffusion &                         & 3.54 & 205.3 & 0.56 \\
        VDM++          &                           & 2.12 & 267.7 & -- \\
        SiD2           &                           & \textbf{1.38} & -- & -- \\
        DiffiT         &                           & 1.73 & 276.5 & 0.62 \\
        \midrule
        JetFormer      & AR+Flow    & 6.64 & --  & 0.56 \\
        RCG            & MAGE                          & 2.15 & 253.4 & 0.53 \\
        \midrule
        FGM            & \multirow{2}{*}{Fractal}    & 6.15 & 348.9 & 0.46 \\
        \textbf{FGM-HD(Ours)} &                    & 6.21 & \textbf{367.1} & \textbf{0.64} \\
        \bottomrule
    \end{tabular}
    \caption{Quantitative evaluation of pixel-level generative models on ImageNet $256 \times 256$. }
    \label{tab:model-comparison}
\end{table}

\begin{table}[t]
    \renewcommand{\arraystretch}{1.1}
    \setlength{\tabcolsep}{5pt}
    \small
     \begin{tabular}{lcccc}
    \toprule
    \textbf{Variant} & \textbf{FID} $\downarrow$ & \textbf{IS} $\uparrow$ & \textbf{Recall} $\uparrow$ & \textbf{LPIPS} $\uparrow$ \\
    \midrule
    FGM (baseline)    & 6.15 & 348.9 & 0.46 & 0.64 \\
    + Fixed HD Loss only       & 6.22 & 333.7 & 0.47 & 0.65 \\
    + MMDS only           & \textbf{6.04} & 361.7 & 0.51 & 0.69 \\
    + HD Sampling only      & 6.78 & 357.9 & 0.58 & 0.73 \\
    + MMDS \& Sampling    & 6.21 & \textbf{367.4} & \textbf{0.64} & \textbf{0.76} \\
    \bottomrule
    \end{tabular}
    \caption{Ablation study of HD-based components. Each component contributes independently to the overall performance, with the combination of MMDS and Sampling Strategy achieving the best balance between diversity and visual quality.}
    \label{tab:hd_ablation}
\end{table}

\subsection{Evaluation Metrics}

To comprehensively evaluate our HD-based strategy effectiveness, we adopted four widely used metrics to jointly assess the diversity and quality of the generated results. We used the \textit{Fréchet inception distance (FID)}~\cite{heusel2017gans} to measure distributional similarity to real images, and \textit{Inception Score (IS)}~\cite{salimans2016improved} to capture both semantic clarity and variety. To assess structural diversity, we used \textit{Recall}~\cite{kynkaanniemi2019improved}, which evaluates the coverage of the target distribution, and \textit{LPIPS}~\cite{zhang2018unreasonable}, which measures the perceptual difference between generated image pairs, providing a direct evaluation of perceptual diversity. These metrics offer a holistic evaluation of the HD loss and HD sampling method's effect on generative quality and diversity.

\subsection{Comparison of HD Estimation Methods}

To assess the accuracy and computational efficiency of different HD estimation techniques, we compared several widely used classical methods and deep learning approaches. Specifically, we evaluated \textit{box counting}, \textit{power spectrum}~\cite{pentland1984fractal}, \textit{perimeter-area scaling}~\cite{russ1994fractal}, and \textit{sandbox method}~\cite{bandt1991deterministic}, as well as deep convolutional regression models based on \textit{ResNet152} and our proposed \textit{multi-scale convolutional network}. All methods were tested on our proposed fractal dataset. Table~\ref{tab:hd-method-comparison} presents each method's average estimation error and runtime per image.

The results indicate that box counting remains the most precise technique partially because it was used to label a part of the dataset. In contrast, the other methods demonstrated higher error and longer runtime. Our multi-scale convolutional network achieves comparable accuracy with faster inference. In practice, especially when using HD to improve output diversity, computational efficiency is often more important than maximum precision. Therefore, a moderate reduction in accuracy is acceptable if the estimates remain consistent.

\begin{figure}[t]
    \centering
    \includegraphics[width=\linewidth]{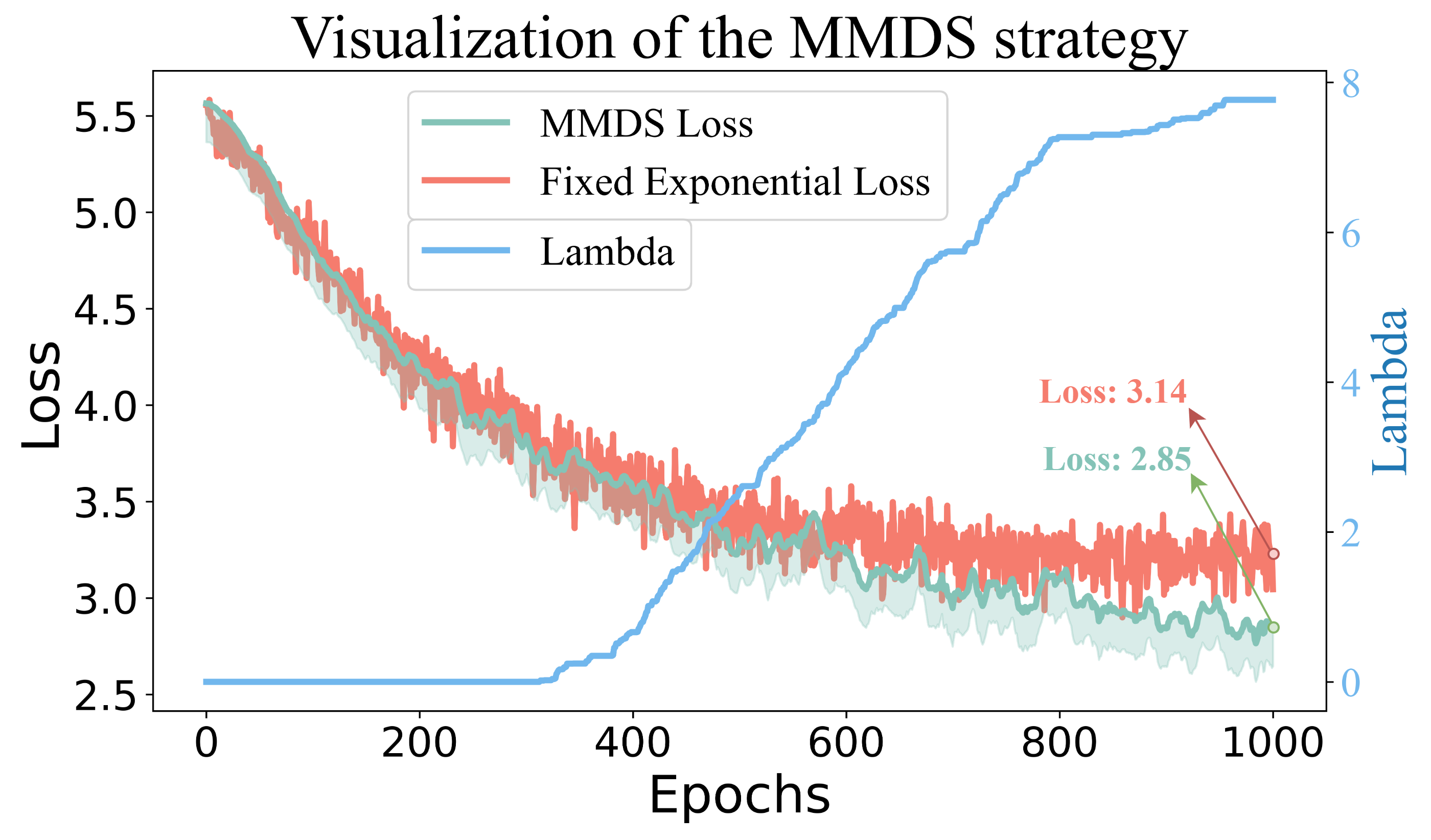}
    \caption{
    Visualization of the MMDS strategy. The blue curve shows the variation of $\lambda$ optimized by MMDS, the red curve represents the fixed exponential loss (final value: 3.14), and the green curve shows the MMDS loss (final value: 2.85). MMDS Strategy leads to a smoother and lower loss trajectory during training.
    }
    \label{fig:lambda_loss_plot}
\end{figure}

\begin{figure*}[ht]
    \centering
    \includegraphics[width=\textwidth]{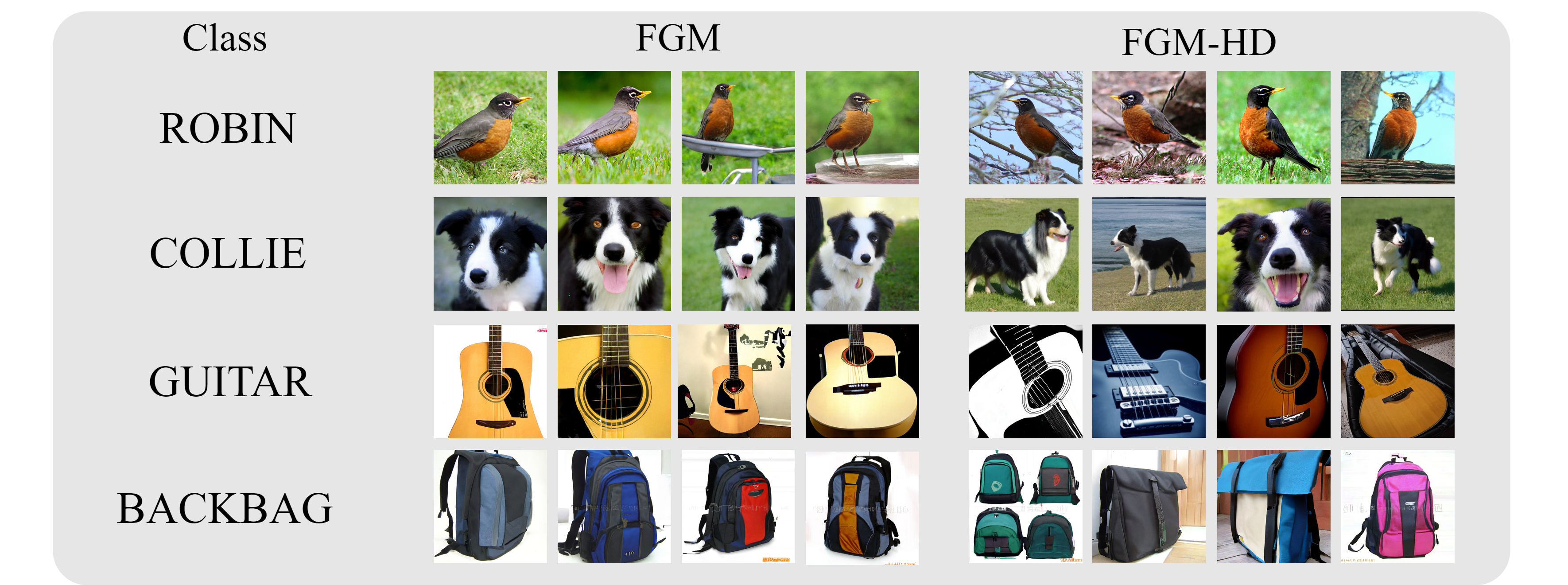}
    \caption{Comparison of generated $256 \times 256$ images between vanilla FGM and FGM-HD across representative ImageNet classes. }

    \label{fig:qualitative_samples}
\end{figure*}

\subsection{Effectiveness of Monotonic Momentum-Driven Scheduling (MMDS)}

To validate our MMDS strategy's effectiveness, we visualized the evolution of $\lambda(t)$ and the corresponding training loss in Figure~\ref{fig:lambda_loss_plot}. The blue curve depicts the momentum-driven schedule, which begins increasing gradually at epoch 300 and plateaus beyond 800. The red and green curves represent the training losses under two different $\lambda(t)$ schedules: the red curve corresponds to a fixed exponential schedule end up to 3.14, while the green curve illustrates our MMDS strategy end up to 2.85. Training was stopped after 1,000 epochs because further training did not significantly improve either quality or diversity. Since MMDS involves a monotonic increase in HD loss, this stopping criterion prevents overfitting by ensuring a balanced focus on visual quality and diversity. The optimal balance between these factors was reached after 1,000 epochs, allowing efficient convergence.

Compared to the fixed exponential schedule, the momentum-driven $\lambda(t)$ yields a significantly smoother and lower loss trajectory. Despite the increasing contribution of HD loss over time, the smoothed loss under dynamic weighting exhibits consistent and stable descent. This indicates that MMDS effectively introduces structural diversity without disrupting training convergence. These results demonstrate that the proposed scheduling mechanism successfully balances structural complexity and optimization stability in generative modeling.

\subsection{HD Sampling Threshold}

To evaluate the of HD sampling effect, we applied varying HD thresholds to the generated samples and monitored performance using FID, IS, Recall, and LPIPS, as shown in Figure~\ref{fig:metrics_hd_combined}. As the threshold increases from 1.1 to around 1.7, FID remains low around a threshold of 1.55 before rising sharply, while IS improves moderately but declines beyond 1.6. Recall steadily increases until 1.7 before dropping. LPIPS, which measures perceptual similarity, increases with the threshold, reflecting improved image quality. However, excessively high values may indicate perceptual distortions, suggesting a trade-off between quality and diversity.

These results suggest that HD thresholds between 1.55 and 1.60 provide an effective balance, preserving sample quality while enhancing structural diversity. This demonstrates the applicability of HD sampling as a targeted post-processing strategy.

\subsection{Comparison with Baseline Generative Models}

To evaluate our proposed HD-based enhancement's effectiveness during the challenging pixel-by-pixel image generation task, we compared several variants of our FGM-HD framework with representative baseline generative models, including \textit{GANs}~\cite{brock2018large, sauer2022stylegan, kang2023scaling}, \textit{diffusion models}~\cite{dhariwal2021diffusion, kingma2021variational, hoogeboom2024simpler, hatamizadeh2024diffit}, and other novel methods~\cite{li2024return, tschannen2024jetformer}. The results are summarized in Table~\ref{tab:model-comparison}.

The experimental results demonstrate that incorporating HD loss during training and HD sampling during inference significantly improves the diversity of the generated results. Specifically, HD loss encourages higher structural complexity, which leads to improved recall and perceptual variation. HD sampling, as an effective post-processing mechanism, further enhances diversity without altering the model architecture. Overall, this combination achieves a 39\% improvement in diversity metrics, while maintaining image quality, as summarized in Table~\ref{tab:hd_ablation}.

\subsection{Qualitative Analysis}
To qualitatively demonstrate our method's capacity to generate diverse and high-quality images, Figure~\ref{fig:qualitative_samples} compares images generated by vanilla FGM (left) and FGM-HD (right). The images span a wide range of semantic categories and exhibit rich structural variation, with noticeable improvements in object shape, color, pose, and background after incorporating HD. These results visually support the improvements in diversity metrics, highlighting the model's ability to synthesize realistic, class-consistent images with enhanced visual diversity. All the images are generated at a resolution of $256 \times 256$ in a pixel-by-pixel manner.


\section{Conclusion}

We proposed a novel HD-guided framework to enhance the structural diversity of FGMs without compromising visual quality. During training, we incorporated HD-based loss with MMDS strategy, which dynamically adjusts the influence of HD loss, ensuring an optimal balance between image quality and structural diversity. During inference, we applied HD-guided rejection sampling to retain only geometrically richer outputs, further promoting diversity. Additionally, we introduced an efficient and learnable HD estimation method that directly predicts HD from image embeddings, significantly improving computational efficiency and accuracy. Our approach demonstrated a 39\% improvement in generation diversity compared to vanilla FGM while maintaining competitive image quality. By introducing HD into the FGM framework, we provide a principled method to enhance diversity without sacrificing quality. Furthermore, the MMDS strategy offers a generalizable optimization technique for hybrid-loss models. Future work includes applying this framework to conditional and multi-modal generative models, and developing perceptually aligned HD estimation methods for large-scale image synthesis.

\section{Acknowledgments}

This work was supported in part by the National Major Scientific Instruments and Equipments Development Project of National Natural Science Foundation of China under Grant 62427820, in part by the Science Fund for Creative Research Groups of Sichuan Province Natural Science Foundation under Grant 2024NSFTD003, in part by the Fundamental Research Funds for the Central Universities under Grant 1082204112364, in part by the Digital Media Art, Key Laboratory of Sichuan Province, Sichuan Conservatory of Music(Grant No. 22DMAKL04). Numerical computations were supported by Chengdu Haiguang Integrated Circuit Design Co., ltd. with HYGON K100AI DCU units.

\bibliography{aaai2026}

@article{goodfellow2014generative,
  title={Generative adversarial nets},
  author={Goodfellow, Ian J and Pouget-Abadie, Jean and Mirza, Mehdi and Xu, Bing and Warde-Farley, David and Ozair, Sherjil and Courville, Aaron and Bengio, Yoshua},
  journal={Advances in neural information processing systems},
  volume={27},
  year={2014}
}

@article{wiatrak2019stabilizing,
  title={Stabilizing generative adversarial networks: A survey},
  author={Wiatrak, Maciej and Albrecht, Stefano V and Nystrom, Andrew},
  journal={arXiv preprint arXiv:1910.00927},
  year={2019}
}

@misc{kingma2013auto,
  title={Auto-encoding variational bayes},
  author={Kingma, Diederik P and Welling, Max and others},
  year={2013},
  publisher={Banff, Canada}
}

@article{van2017neural,
  title={Neural discrete representation learning},
  author={Van Den Oord, Aaron and Vinyals, Oriol and others},
  journal={Advances in neural information processing systems},
  volume={30},
  year={2017}
}

@article{ho2020denoising,
  title={Denoising diffusion probabilistic models},
  author={Ho, Jonathan and Jain, Ajay and Abbeel, Pieter},
  journal={Advances in neural information processing systems},
  volume={33},
  pages={6840--6851},
  year={2020}
}

@inproceedings{rombach2022high,
  title={High-resolution image synthesis with latent diffusion models},
  author={Rombach, Robin and Blattmann, Andreas and Lorenz, Dominik and Esser, Patrick and Ommer, Bj{\"o}rn},
  booktitle={Proceedings of the IEEE/CVF conference on computer vision and pattern recognition},
  pages={10684--10695},
  year={2022}
}

@article{shen2025efficient,
  title={Efficient Diffusion Models: A Survey},
  author={Shen, Hui and Zhang, Jingxuan and Xiong, Boning and Hu, Rui and Chen, Shoufa and Wan, Zhongwei and Wang, Xin and Zhang, Yu and Gong, Zixuan and Bao, Guangyin and others},
  journal={arXiv preprint arXiv:2502.06805},
  year={2025}
}

@article{dhariwal2021diffusion,
  title={Diffusion models beat gans on image synthesis},
  author={Dhariwal, Prafulla and Nichol, Alexander},
  journal={Advances in neural information processing systems},
  volume={34},
  pages={8780--8794},
  year={2021}
}

@article{li2025fractal,
  title={Fractal generative models},
  author={Li, Tianhong and Sun, Qinyi and Fan, Lijie and He, Kaiming},
  journal={arXiv preprint arXiv:2502.17437},
  year={2025}
}

@article{hausdorff1918dimension,
  title={Dimension und {\"a}u{\ss}eres Ma{\ss}},
  author={Hausdorff, Felix},
  journal={Mathematische Annalen},
  volume={79},
  number={1},
  pages={157--179},
  year={1918},
  publisher={Springer}
}

@book{barnsley2014fractals,
  title={Fractals everywhere},
  author={Barnsley, Michael F},
  year={2014},
  publisher={Academic press}
}

@book{prusinkiewicz2012algorithmic,
  title={The algorithmic beauty of plants},
  author={Prusinkiewicz, Przemyslaw and Lindenmayer, Aristid},
  year={2012},
  publisher={Springer Science \& Business Media}
}

@article{khrulkov2019universality,
  title={Universality theorems for generative models},
  author={Khrulkov, Valentin and Oseledets, Ivan},
  journal={arXiv preprint arXiv:1905.11520},
  year={2019}
}

@article{simsekli2020hausdorff,
  title={Hausdorff dimension, heavy tails, and generalization in neural networks},
  author={Simsekli, Umut and Sener, Ozan and Deligiannidis, George and Erdogdu, Murat A},
  journal={Advances in Neural Information Processing Systems},
  volume={33},
  pages={5138--5151},
  year={2020}
}

@article{li2021hausdorff,
  title={Hausdorff GAN: Improving GAN generation quality with Hausdorff metric},
  author={Li, Wei and Liang, Zhixuan and Ma, Ping and Wang, Ruobei and Cui, Xiaohui and Chen, Ping},
  journal={IEEE Transactions on Cybernetics},
  volume={52},
  number={10},
  pages={10407--10419},
  year={2021},
  publisher={IEEE}
}

@article{brock2018large,
  title={Large scale GAN training for high fidelity natural image synthesis},
  author={Brock, Andrew and Donahue, Jeff and Simonyan, Karen},
  journal={arXiv preprint arXiv:1809.11096},
  year={2018}
}

@inproceedings{kang2023scaling,
  title={Scaling up gans for text-to-image synthesis},
  author={Kang, Minguk and Zhu, Jun-Yan and Zhang, Richard and Park, Jaesik and Shechtman, Eli and Paris, Sylvain and Park, Taesung},
  booktitle={Proceedings of the IEEE/CVF conference on computer vision and pattern recognition},
  pages={10124--10134},
  year={2023}
}

@inproceedings{sauer2022stylegan,
  title={Stylegan-xl: Scaling stylegan to large diverse datasets},
  author={Sauer, Axel and Schwarz, Katja and Geiger, Andreas},
  booktitle={ACM SIGGRAPH 2022 conference proceedings},
  pages={1--10},
  year={2022}
}

@article{kingma2021variational,
  title={Variational diffusion models},
  author={Kingma, Diederik and Salimans, Tim and Poole, Ben and Ho, Jonathan},
  journal={Advances in neural information processing systems},
  volume={34},
  pages={21696--21707},
  year={2021}
}

@article{hoogeboom2024simpler,
  title={Simpler diffusion (sid2): 1.5 fid on imagenet512 with pixel-space diffusion},
  author={Hoogeboom, Emiel and Mensink, Thomas and Heek, Jonathan and Lamerigts, Kay and Gao, Ruiqi and Salimans, Tim},
  journal={arXiv preprint arXiv:2410.19324},
  year={2024}
}

@article{tschannen2024jetformer,
  title={JetFormer: An autoregressive generative model of raw images and text},
  author={Tschannen, Michael and Pinto, Andr{\'e} Susano and Kolesnikov, Alexander},
  journal={arXiv preprint arXiv:2411.19722},
  year={2024}
}

@article{heusel2017gans,
  title={Gans trained by a two time-scale update rule converge to a local nash equilibrium},
  author={Heusel, Martin and Ramsauer, Hubert and Unterthiner, Thomas and Nessler, Bernhard and Hochreiter, Sepp},
  journal={Advances in neural information processing systems},
  volume={30},
  year={2017}
}

@inproceedings{zhang2018unreasonable,
  title={The unreasonable effectiveness of deep features as a perceptual metric},
  author={Zhang, Richard and Isola, Phillip and Efros, Alexei A and Shechtman, Eli and Wang, Oliver},
  booktitle={Proceedings of the IEEE conference on computer vision and pattern recognition},
  pages={586--595},
  year={2018}
}

@article{salimans2016improved,
  title={Improved techniques for training gans},
  author={Salimans, Tim and Goodfellow, Ian and Zaremba, Wojciech and Cheung, Vicki and Radford, Alec and Chen, Xi},
  journal={Advances in neural information processing systems},
  volume={29},
  year={2016}
}

@article{kynkaanniemi2019improved,
  title={Improved precision and recall metric for assessing generative models},
  author={Kynk{\"a}{\"a}nniemi, Tuomas and Karras, Tero and Laine, Samuli and Lehtinen, Jaakko and Aila, Timo},
  journal={Advances in neural information processing systems},
  volume={32},
  year={2019}
}

@inproceedings{van2016pixel,
  title={Pixel recurrent neural networks},
  author={Van Den Oord, A{\"a}ron and Kalchbrenner, Nal and Kavukcuoglu, Koray},
  booktitle={International conference on machine learning},
  pages={1747--1756},
  year={2016},
  organization={PMLR}
}

@article{kingma2018glow,
  title={Glow: Generative flow with invertible 1x1 convolutions},
  author={Kingma, Durk P and Dhariwal, Prafulla},
  journal={Advances in neural information processing systems},
  volume={31},
  year={2018}
}

@article{you2022locally,
  title={Locally hierarchical auto-regressive modeling for image generation},
  author={You, Tackgeun and Kim, Saehoon and Kim, Chiheon and Lee, Doyup and Han, Bohyung},
  journal={Advances in Neural Information Processing Systems},
  volume={35},
  pages={16360--16372},
  year={2022}
}

@inproceedings{deng2009imagenet,
  title={Imagenet: A large-scale hierarchical image database},
  author={Deng, Jia and Dong, Wei and Socher, Richard and Li, Li-Jia and Li, Kai and Fei-Fei, Li},
  booktitle={2009 IEEE conference on computer vision and pattern recognition},
  pages={248--255},
  year={2009},
  organization={Ieee}
}

@article{li2024autoregressive,
  title={Autoregressive image generation without vector quantization},
  author={Li, Tianhong and Tian, Yonglong and Li, He and Deng, Mingyang and He, Kaiming},
  journal={Advances in Neural Information Processing Systems},
  volume={37},
  pages={56424--56445},
  year={2024}
}

@article{mandelbrot1983fractal,
  title={The fractal geometry of nature/Revised and enlarged edition},
  author={Mandelbrot, Benoit B},
  journal={New York},
  year={1983}
}

@article{gneiting2012estimators,
  title={Estimators of fractal dimension: Assessing the roughness of time series and spatial data},
  author={Gneiting, Tilmann and {\v{S}}ev{\v{c}}{\'\i}kov{\'a}, Hana and Percival, Donald B},
  journal={Statistical Science},
  pages={247--277},
  year={2012},
  publisher={JSTOR}
}

@incollection{napolitano2012fractal,
  title={Fractal dimension estimation methods for biomedical images},
  author={Napolitano, Antonio and Ungania, Sara and Cannata, Vittorio},
  booktitle={MATLAB-A Fundamental Tool for Scientific Computing and Engineering Applications-Volume 3},
  year={2012},
  publisher={IntechOpen}
}

@article{lecun1989backpropagation,
  title={Backpropagation applied to handwritten zip code recognition},
  author={LeCun, Yann and Boser, Bernhard and Denker, John S and Henderson, Donnie and Howard, Richard E and Hubbard, Wayne and Jackel, Lawrence D},
  journal={Neural computation},
  volume={1},
  number={4},
  pages={541--551},
  year={1989},
  publisher={MIT Press}
}

@inproceedings{he2016deep,
  title={Deep residual learning for image recognition},
  author={He, Kaiming and Zhang, Xiangyu and Ren, Shaoqing and Sun, Jian},
  booktitle={Proceedings of the IEEE conference on computer vision and pattern recognition},
  pages={770--778},
  year={2016}
}

@article{valle2022characterization,
  title={Characterization of fractal basins using deep convolutional neural networks},
  author={Valle, David and Wagemakers, Alexandre and Daza, Alvar and Sanju{\'a}n, Miguel AF},
  journal={International Journal of Bifurcation and Chaos},
  volume={32},
  number={13},
  pages={2250200},
  year={2022},
  publisher={World Scientific}
}

@article{pentland1984fractal,
  title={Fractal-based description of natural scenes},
  author={Pentland, Alex P},
  journal={IEEE transactions on pattern analysis and machine intelligence},
  number={6},
  pages={661--674},
  year={1984},
  publisher={IEEE}
}

@book{russ1994fractal,
  title={Fractal surfaces},
  author={Russ, John C},
  year={1994},
  publisher={Springer Science \& Business Media}
}

@article{bandt1991deterministic,
  title={Deterministic fractals and fractal measures},
  author={Bandt, Christoph},
  year={1991},
  publisher={Universit{\`a} degli Studi di Trieste. Dipartimento di Scienze Matematiche}
}

@article{robbins1951stochastic,
  title={A stochastic approximation method},
  author={Robbins, Herbert and Monro, Sutton},
  journal={The Annals of Mathematical Statistics},
  volume={22},
  number={3},
  pages={400--407},
  year={1951},
  publisher={Institute of Mathematical Statistics}
}

@inproceedings{kataoka2020pre,
  title={Pre-training without natural images},
  author={Kataoka, Hirokatsu and Okayasu, Kazushige and Matsumoto, Asato and Yamagata, Eisuke and Yamada, Ryosuke and Inoue, Nakamasa and Nakamura, Akio and Satoh, Yutaka},
  booktitle={Proceedings of the Asian Conference on Computer Vision},
  year={2020}
}

@article{werbos2002backpropagation,
  title={Backpropagation through time: what it does and how to do it},
  author={Werbos, Paul J},
  journal={Proceedings of the IEEE},
  volume={78},
  number={10},
  pages={1550--1560},
  year={2002},
  publisher={IEEE}
}

@article{li2024return,
  title={Return of unconditional generation: A self-supervised representation generation method},
  author={Li, Tianhong and Katabi, Dina and He, Kaiming},
  journal={Advances in Neural Information Processing Systems},
  volume={37},
  pages={125441--125468},
  year={2024}
}

@inproceedings{hatamizadeh2024diffit,
  title={Diffit: Diffusion vision transformers for image generation},
  author={Hatamizadeh, Ali and Song, Jiaming and Liu, Guilin and Kautz, Jan and Vahdat, Arash},
  booktitle={European Conference on Computer Vision},
  pages={37--55},
  year={2024},
  organization={Springer}
}

@inproceedings{lan2024generative,
  title={Generative model perception rectification algorithm for trade-off between diversity and quality},
  author={Lan, Guipeng and Xiao, Shuai and Yang, Jiachen and Wen, Jiabao},
  booktitle={Proceedings of the AAAI conference on artificial intelligence},
  volume={38},
  number={12},
  pages={13328--13336},
  year={2024}
}

@article{li2021combination,
  title={Combination of certainty and uncertainty: Using FusionGAN to create abstract paintings},
  author={Li, Mao and Lv, Jiancheng and Tang, Chenwei and Wang, Jian and Lai, Zhichen and Huang, Youcheng},
  journal={Neural Networks},
  volume={144},
  pages={443--454},
  year={2021},
  publisher={Elsevier}
}

\clearpage  
\appendix
\section{Appendix A: HD Estimation Network}

We provide the architectural specifications and training configuration for our multi-scale convolutional network which directly estimates HD from image embeddings.

\subsection{Network Architecture}

Our HD estimation model builds on the ResNet152 backbone, known for its strong feature extraction capabilities. To improve the model's HD estimation ability, we replaced the last two layers of ResNet152 with a multi-scale convolutional module.

This module uses parallel convolutional branches with $3 \times 3$, $5 \times 5$, and $7 \times 7$ kernel sizes applied to the shared intermediate feature maps from ResNet152's last block. This multi-scale approach captures spatial information at different scales, allowing for better extraction of the structural details crucial for HD estimation.

Replacing the final layers, which typically focus on classification, with our custom module enables direct HD calculation, retaining high-level feature information throughout the network. We experimented with various strategies and discovered that replacing the fourth and fifth layers offers the best balance between efficiency and accuracy, leading to superior HD estimation performance.

\begin{table}[h]
	\centering
	\small
	\renewcommand{\arraystretch}{1.1}
	\setlength{\tabcolsep}{4pt}
	\begin{tabular}{llcc}
		\toprule
		\textbf{Stage} & \textbf{Component} & \textbf{Kernel / Stride} & \textbf{Channels} \\
		\midrule
		Stem        & Conv + BN + ReLU    & $7\times7$ / 2     & 64 \\
		& MaxPool             & $3\times3$ / 2     & 64 \\
		Conv2\_x    & Bottleneck-1        & $1$ / 1, $3$ / 1, $1$ / 1 & 256 \\
		& Bottleneck-2        & $1$ / 1, $3$ / 1, $1$ / 1 & 256 \\
		& Bottleneck-3        & $1$ / 1, $3$ / 1, $1$ / 1 & 256 \\
		Multi-Scale & Conv$_1$            & $3\times3$ / 1     & 64 \\
		& Conv$_2$            & $5\times5$ / 1     & 64 \\
		& Conv$_3$            & $7\times7$ / 1     & 64 \\
		& Concat + GAP        & --                 & 192 \\
		Output      & Fully Connected     & --                 & 1 \\
		\bottomrule
	\end{tabular}
	\caption{HD estimation network architecture.}
	\label{tab:hd_net_arch}
\end{table}

\subsection{Multi-scale Replacement Experiments}

We performed a series of experiments by replacing different convolutional layers in the ResNet152 architecture to improve the efficiency and accuracy of HD estimation. The original ResNet152 consists of five stages, where each stage progressively extracts deeper features through multiple convolutions, with each successive stage reducing the image size.

Thus, we conducted experiments by replacing various stages of the convolutional layers with a multi-scale convolution module ($3 \times 3$, $5 \times 5$, $7 \times 7$ kernels). Additionally, we performed an ablation study on the selection of kernel sizes in the multi-scale convolutional module in Table~\ref{tab:ablation_study}. The results demonstrated that replacing the \textbf{Stages 4 and 5} layers provided the best balance between HD estimation efficiency and accuracy in Table~\ref{tab:performance_comparison}. While replacing more stages of ResNet152 reduced time consumption, it also resulted in a loss of deeper convolutional image embeddings, significantly impacting accuracy. Furthermore, the choice of kernel size played a crucial role in performance. Combining $3 \times 3$, $5 \times 5$, and $7 \times 7$ kernels yielded the best results, capturing fine-grained details and larger, global structures.

The modified network, with only Stage 4 and 5 replaced, and the optimal kernel combination, extracted structural features more effectively, leading to improved HD prediction with optimal performance.

\begin{table}[h]
	\centering
	\renewcommand{\arraystretch}{1.2}
	\setlength{\tabcolsep}{6pt}
	\begin{tabular}{|l|c|c|}
		\hline
		\textbf{Kernel Size} & \textbf{Time (s)} & \textbf{Loss} \\ \hline
		(a) & \textbf{0.30} & 0.01 \\ \hline
		(b) & 0.33 & 0.009 \\ \hline
		(c) & 0.38 & 0.009 \\ \hline
		(a, b) & 0.31 & 0.007 \\ \hline
		(b, c) & 0.34 & 0.007 \\ \hline
		(a, b, c) & 0.32 & \textbf{0.005} \\ \hline
	\end{tabular}
	\caption{Ablation study on multi-scale convolutions for HD estimation. Kernels: (a) $3 \times 3$, (b) $5 \times 5$, (c) $7 \times 7$.}
	\label{tab:ablation_study}
\end{table}

\begin{table}[h]
	\centering
	
	\small
	\renewcommand{\arraystretch}{1.1}
	\setlength{\tabcolsep}{6pt}
	\begin{tabular}{|l|c|c|}
		\hline
		\textbf{Model Architecture} & \textbf{Time(s)} & \textbf{Accuracy Error} \\ \hline
		Full ResNet152 & 0.40 & 0.012 \\ \hline
		Replacement for Stage 5 & 0.37 & 0.008 \\ \hline
		Replacement for Stages 4-5 & 0.32 & \textbf{0.005} \\ \hline
		Replacement for Stages 3-5 & 0.29 & 0.020 \\ \hline
		Replacement for Stages 2-5 & \textbf{0.27} & 0.037 \\ \hline
	\end{tabular}
	\caption{Performance comparison of different ResNet152 architectures for HD estimation.}
	\label{tab:performance_comparison}
\end{table}

\subsection{Training Details}
The model was trained using the mean squared error (MSE) loss between the predicted and ground-truth HD values. Optimization was performed using the Adam optimizer with an initial learning rate of $1e^{-4}$, batch size of 32, and training for 300 epochs. All input images were resized to $224 \times 224$ and normalized to $[0, 1]$.

\begin{figure*}[t]
	\centering
	\includegraphics[width=0.95\linewidth]{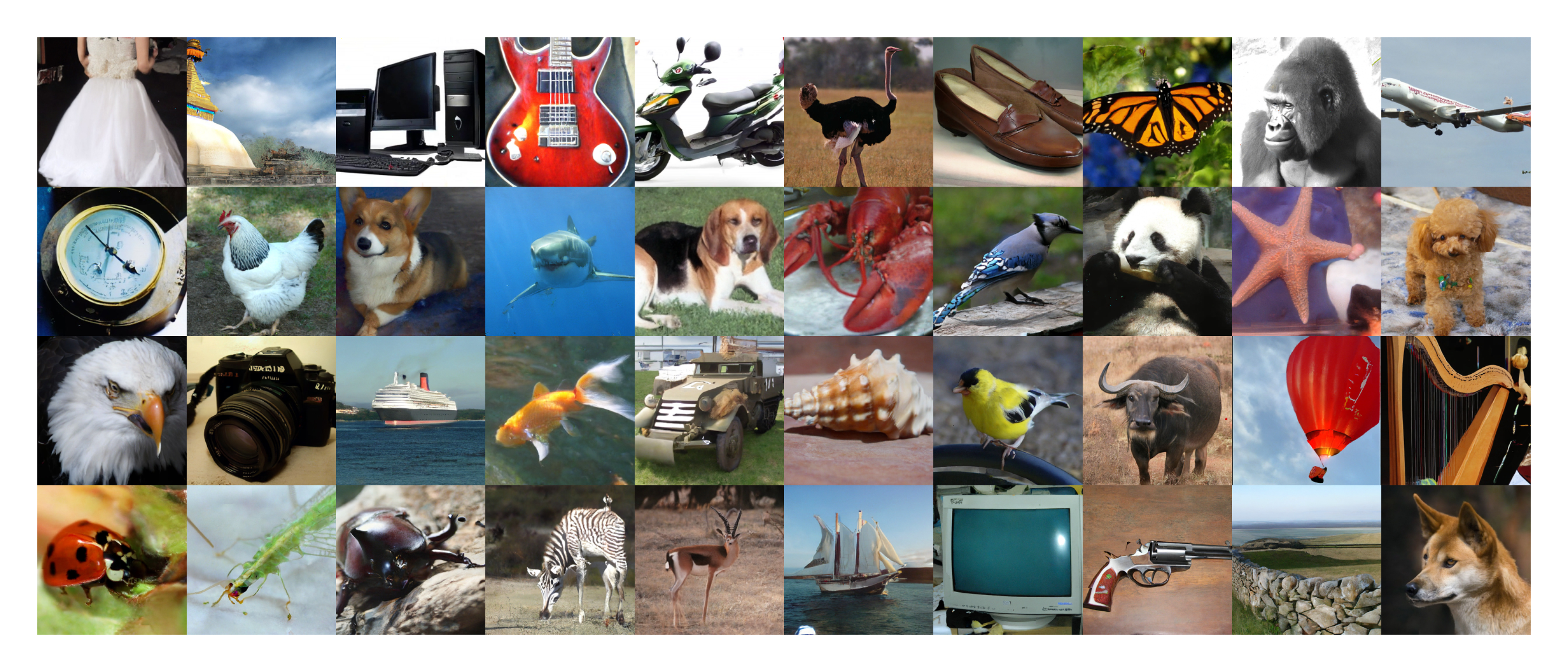}
	\caption{Additional samples generated by FGM-HD across different categories, showcasing structural complexity and visual diversity.}
	\label{fig:appendix_gallery}
\end{figure*}

\vspace{1em}
\section{Appendix B: Training Strategy}

Our HD estimation model's training incorporated the Monotonic MMDS strategy, which gradually introduces the influence of the HD loss over time. This approach ensures the model strikes a balance between visual quality and structural diversity, allowing for efficient convergence without overfitting.

\begin{figure}[t]
	\centering
	\includegraphics[width=\linewidth]{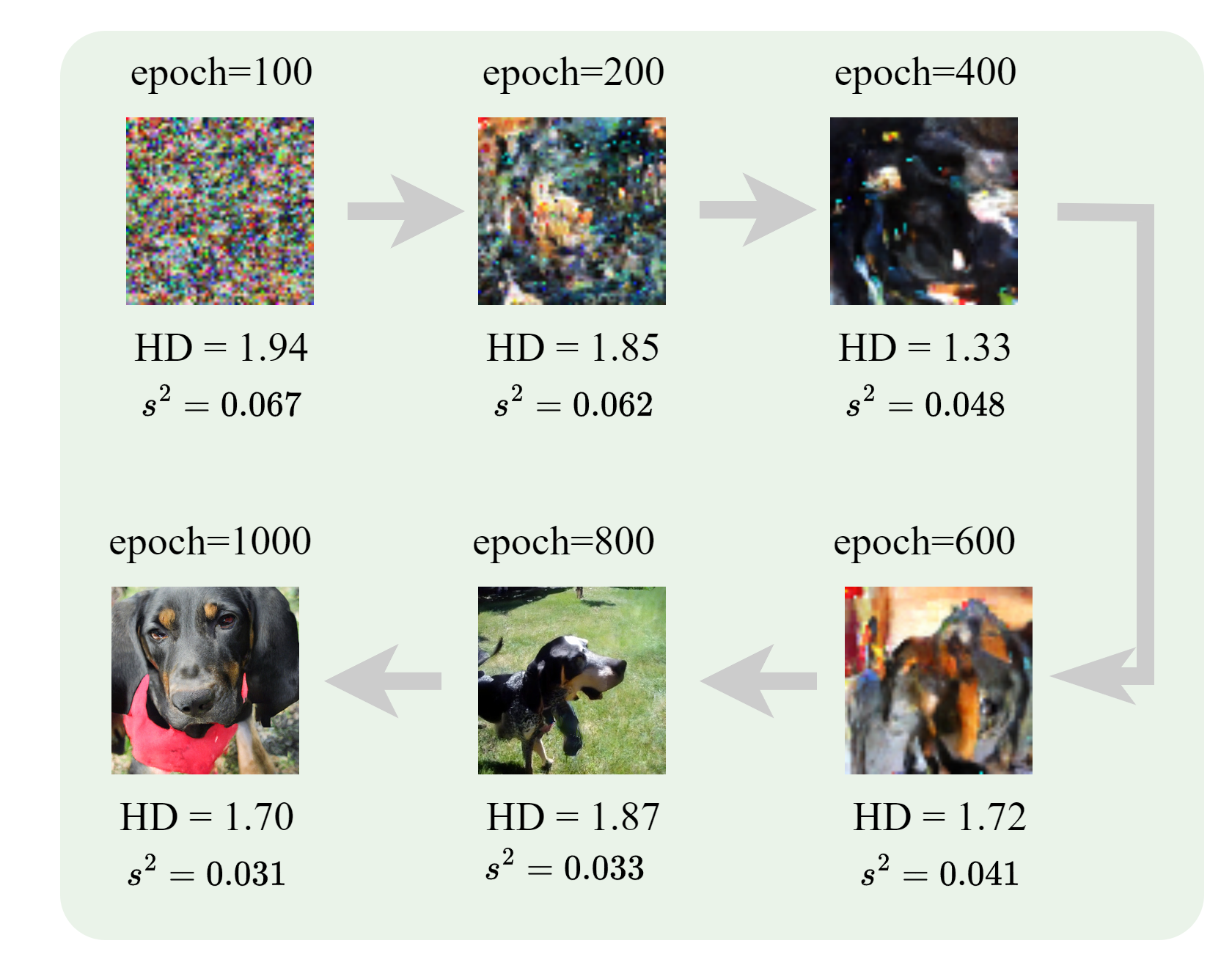}
	\caption{Evolution of image quality and HD variance across training epochs. Early-stage generations are noisy with unstable HD values, while later epochs yield clearer images with more reliable HD estimations.}
	\label{fig:epoch-hd}
\end{figure}

In the early stages of training (e.g., epoch 100–300), the model produced highly noisy outputs with significant variance, as illustrated in Figure~\ref{fig:epoch-hd}. These early-stage images lack coherent structure, resulting in unreliable HD estimations that fluctuate considerably (e.g., HD = 1.94, $s^2$ = 0.067 at epoch 100). Introducing HD loss too early under such conditions could destabilize training and deteriorate image fidelity. As training advances, image quality progressively improved, and HD estimations became more consistent and meaningful (e.g., by epoch 800–1000, HD stabilized around 1.7 with low variance). This justified our strategy of delaying the incorporation of HD loss until the model has reached a more stable generative phase. By gradually increasing the HD loss weight after epoch 300, our MMDS stratrgy enabled a smooth transition toward enhanced structural diversity, without compromising visual quality.

Training was stopped after 1,000 epochs, as further training did not significantly improve the loss or diversity metrics. The MMDS strategy gradually increased the HD loss, and by epoch 1,000, the model achieved the optimal balance between visual quality and diversity. Continuing beyond this point would likely result in diminishing returns, as further adjustments to the HD loss did not yield substantial gains. This stopping criterion prevents overfitting, ensuring the model maintains a good balance and converges efficiently at epoch 1,000, achieving the desired performance. This strategy allows the model to learn effectively and efficiently, focusing on quality and diversity while avoiding unnecessary computational cost from further epochs.

\begin{figure}[t]
	\centering
	\includegraphics[width=0.95\linewidth]{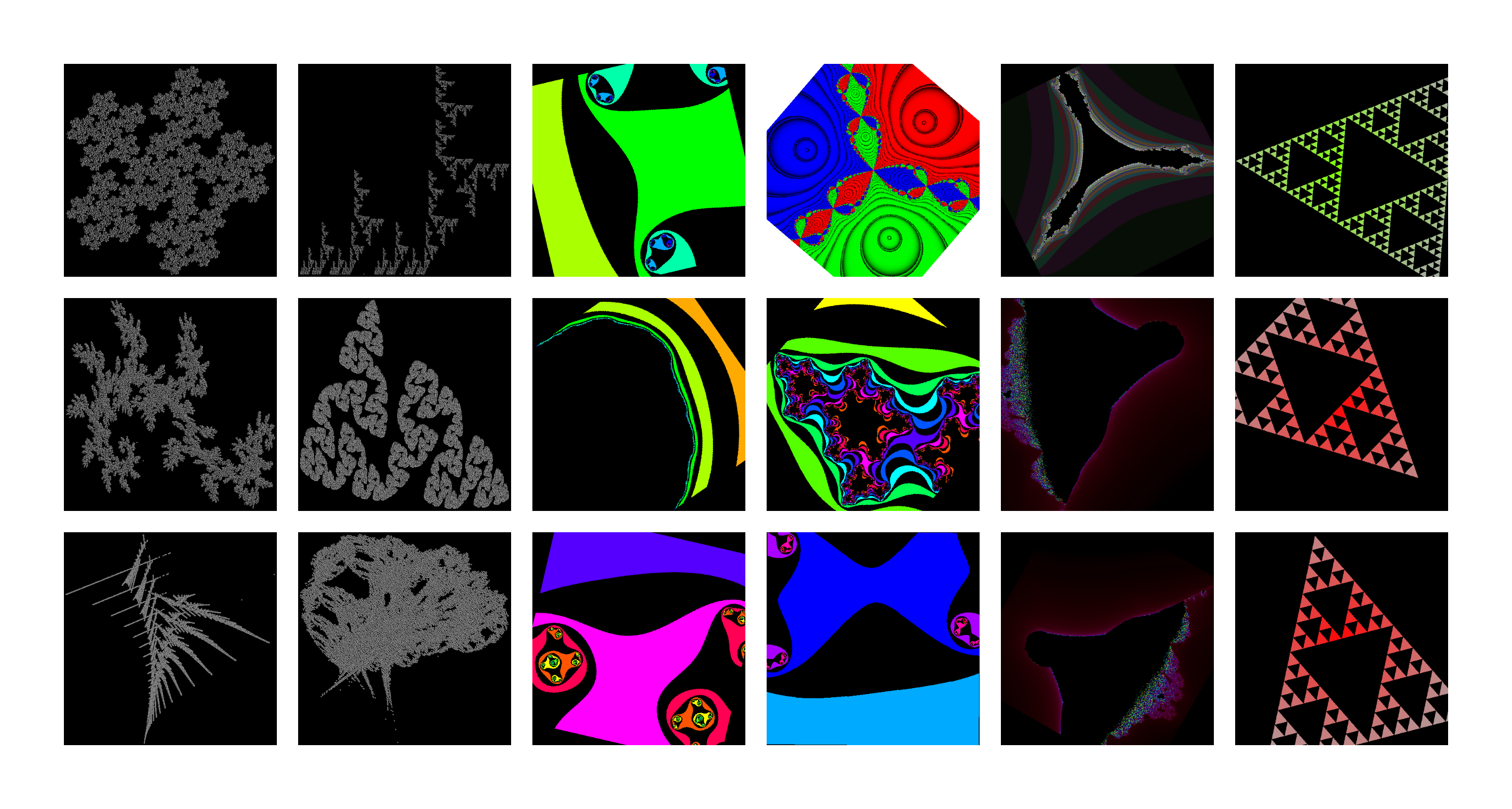}
	\caption{Representative samples from the constructed fractal dataset.}
	\label{fig:fractal_dataset_gallery}
\end{figure}

\vspace{1em}

\section{Appendix C: Datasets}

\subsection{Fractal Dataset Construction}

We constructed a dedicated fractal dataset to benchmark HD estimation and diversity evaluation in a controlled setting, as shown in Figure~\ref{fig:fractal_dataset_gallery}. The dataset contains 1,200 synthetic images with a $256 \times 256$ resolution, consisting of:

\begin{itemize}
	\item \textbf{Canonical fractal images} (1/3): e.g., Sierpinski triangle, Mandelbrotset, and Koch snowflake, with known theoretical HD values.
	\item \textbf{IFS-generated fractals} (1/3): Images generated using the IFS with varying contraction coefficients and affine rules to produce diverse recursive structures.
	\item \textbf{FractalDB samples with HD approximation} (1/3): A curated subset of fractal textures from FractalDB, where HD values are estimated using the box counting method and used as regression targets.
\end{itemize}

\begin{figure}[t]
	\centering
	\includegraphics[width=0.95\linewidth]{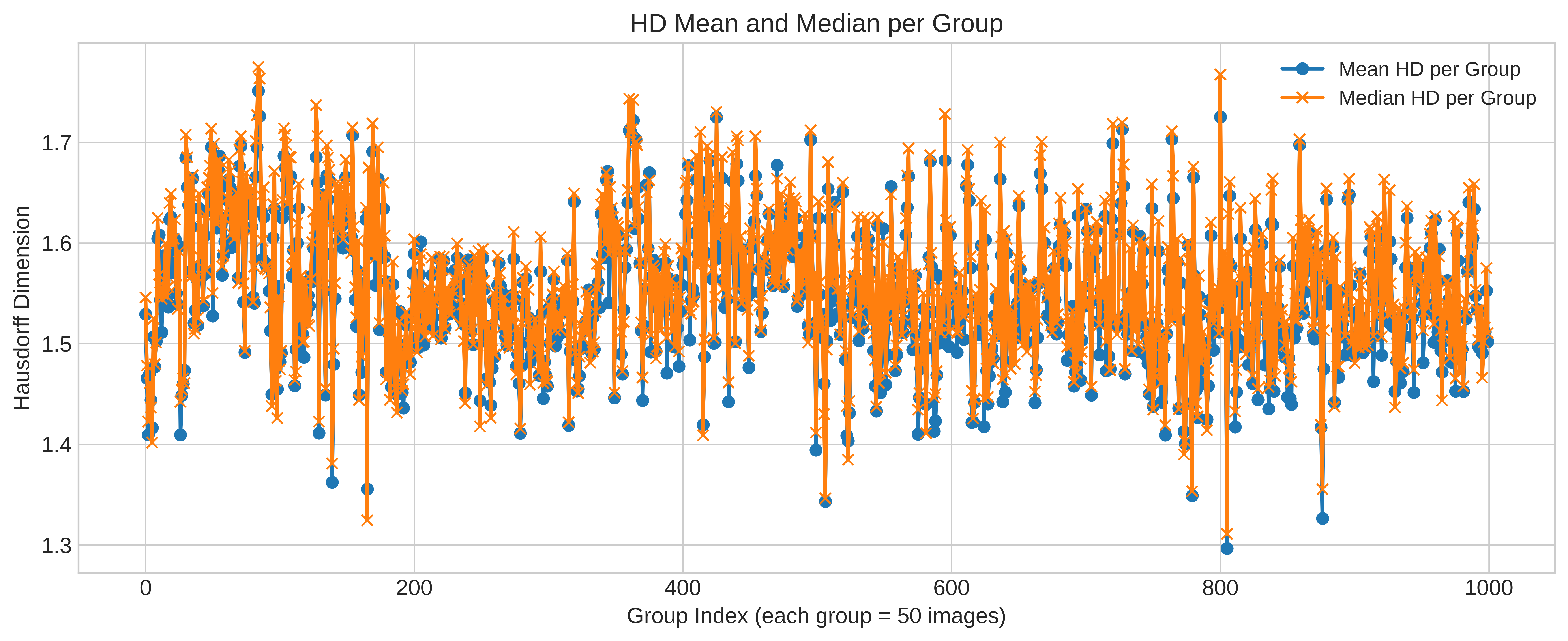}
	\caption{The mean and median HD values from the 1000 ImageNet categories.}
	\label{fig:group}
\end{figure}

\subsection{ImageNet HD Estimation}

In addition, we evaluated the HD estimation on the ImageNet-1000 dataset, where we generated 50 images per category. For each image, we computed the median and mean HD values for each category. This provided a wide range of HD distribution values spanning from 1.1 to 1.9, covering a diverse set of geometric complexities across the dataset.

The HD values computed for ImageNet-1000 images were used as part of the thresholding mechanism for HD-based filtering. These computed HD values served as a preliminary standard for threshold selection, further fine-tuned as described in HD Sampling Threshold section. Furthermore, this dataset enabled the quantitative validation of our HD estimation models and supported HD-supervised training and filtering pipelines. Example images from each category are illustrated in Figure~\ref{fig:group}.

\vspace{1em}
\section{Appendix D: Qualitative Sample Gallery}

To further illustrate our FGM-HD framework's effectiveness, a diverse selection of $256 \times 256$ samples generated using our FGM-HD are presented in Figure~\ref{fig:appendix_gallery}. These examples span a wide range of semantic categories and demonstrate the model’s ability to synthesize visually coherent and structurally richer outputs.

\section{Appendix E: Parameter Selection}

We performed a series of experiments to evaluate the impact of $\mu$ and $\gamma$ on the MMDS strategy shown in Table~\ref{tab:experiment}. The smoothness of the loss curve was quantified using a sliding window approach, which measures the extent of fluctuations in the loss curve. For each experiment, we applied a sliding window technique, moving a window of fixed size over the loss curve to compute a statistic (e.g., mean, variance) for each window position. This allowed us to observe the fluctuations in the loss curve and compute a smoothness score.

The table below shows the results of our experiments, where we observed that $\mu = 0.9$ and $\gamma = 1.0$ provided a good balance between performance (Loss) and stability (Smoothness). 

\begin{table}[ht]
	\centering
	\begin{tabular}{|c|c|c|c|c|}
		\hline
		\textbf{$\mu$} & \textbf{$\gamma$} & \textbf{Loss} & \textbf{Smooth} & \textbf{Convergence} \\
		\hline
		0.9 & 0.5 & 3.09 & 0.041 & $>1000$ \\
		0.9 & 1.0 & 2.85 & 0.047 & 800 \\
		0.9 & 5.0 & 6.22 & 0.088 & 700 \\
		0.8 & 1.0 & 2.98 & 0.045 & 900 \\
		0.95 & 1.0 & 2.83 & 0.062 & 800 \\
		\hline
	\end{tabular}
	\caption{Impact of $\mu$ and $\gamma$ on MMDS strategy performance.}
	\label{tab:experiment}
\end{table}




\end{document}